\documentclass{article}
\usepackage{spconf,amsmath,graphicx,float,stfloats}
\usepackage{amsfonts}
\usepackage{booktabs} 
\usepackage{varwidth} 
\usepackage{graphicx}
\usepackage{cite}
\usepackage{array}

\title{Disentangling the Spatial Structure and Style in Conditional VAE}
\name{Ziye Zhang$^1$, Li Sun$^{1,2,3}$\sthanks{Corresponding author (Email: sunli@ee.ecnu.edu.cn). This work is supported by the the Science and Technology Commission of Shanghai Municipality (No.19511120800), and the Open Project Program of the State Key Laboratory of Mathematical Engineering and Advanced Computing.}, Zhilin Zheng$^1$ and Qingli Li$^1$}
\address{$^1$Shanghai Key Laboratory of Multidimensional Information Processing, \\$^2$Key Laboratory of Advanced Theory and Application in Statistics and Data Science, \\East China Normal University, 200241 Shanghai, China\\$^3$State Key Laboratory of Mathematical Engineering and Advanced Computing, 214125 Wuxi, China}

\begin{document}

\maketitle
\vspace{-0.4pt}
\begin{abstract}
This paper proposes a structure in conditional variation autoencoder (cVAE) to disentangle the latent vector into a spatial structure and a style code, complementary to each other, with the one ($z_s$) being label relevant and the other ($z_u$) irrelevant. 
Different from traditional cVAE, our network maps the condition label into its relevant code $z_s$ through a separated module. Depending on whether the label directly relates to the image spatial structure or not, $z_s$ output from the condition mapping module is used either as the style code with the two spatial dimension of $1\times1$, or as the spatial structure code with a single channel. Based on the input image and its corresponding $z_s$, the encoder provides the posterior distribution close to a common prior regardless of its label, thus $z_u$ sampled from it becomes label irrelevant. The decoder employs $z_s$ and $z_u$ 
by two typical adaptive normalization modules 
to reconstruct the input image. Results on two datasets with different types of labels show the effectiveness of our method. 
\end{abstract}
\vspace{-0.2cm}
\section{Introduction}
\vspace{-0.1cm}
\label{sec:intro}

VAE \cite{rezende2014stochastic} and GAN \cite{goodfellow2014generative} are two powerful tools for image synthesis. In GAN, the generator $G(z)$ aims to mimic the data distribution $p_{\rm{data}}(x)$ with an approximation $p_G(z)$ by mapping the random noise $z$ 
to the image-like data. 
Meanwhile, GAN learns a discriminator $D$ 
to distinguish the samples, either drawn from $p_{\rm{data}}({\bf x})$ or $p_G({\bf z})$. $G$ and $D$ are trained jointly in an adversarial manner. VAE consists of a pair of connected encoder and decoder, with their parameters $\phi$ and $\theta$, respectively. The encoder maps the image $x$ into a posterior distribution $q_\phi(z|x)$ from which the code $z$ is sampled, and the decoder 
$p_\theta(x|z)$, transforms
$z$ back into image domain to reconstruct $x$.
VAE requires $q_\phi(z|x)$ to be simple, \emph{e.g.}, close to the Gaussian prior $N(z|0, I)$ based on the KL divergence. 

Compared to GAN, VAE tends to generate blurry images, since $q_\phi(z|x)$
is too simple to capture the true posterior, known as "posterior collapse". But it is easier to train. While GAN's optimization is unstable, hence many works try to stabilize its training \cite{arjovsky2017wasserstein,gulrajani2017improved,miyato2018spectral}. Moreover, VAE explicitly models each dimension of $z$ as independent Gaussian, so it can disentangle the factors in unsupervised way \cite{higgins2017beta,chen2018isolating}. To fully exploit the advantage from both of them, VAE and GAN can be combined into VAE-GAN \cite{larsen2015autoencoding}, in which the encoder and decoder in VAE forms the generator, and it employs a discriminator to identify the real from the fake image.

Both GAN and VAE can be utilized for conditional generation. The generator in conditional GAN (cGAN) \cite{mirza2014conditional,miyato2018cgans} is usually given the concatenation of a random code $z$ and a conditional label $c$. Its output $G(z,c)$ is required to fulfill the condition. Here $c$ has various forms. It can be a one-hot vector indicating the categories, or a conditional image with its spatial structure. $D$ in cGAN not only evaluates the reality of $G(z,c)$, but also checks its conformity on $c$. Similarly, cVAE \cite{sohn2015learning} gives the label $c$ to both encoder and decoder. The posterior specified by the encoder becomes $q_\phi(z|x,c)$. 
Note that $x$ with different $c$ are mapping to the same prior $N(z;0, I)$, so the regularization term
actually 
prevents $z$ being relevant with $c$ in some extent. Then the decoder $p_\theta(x|z,c)$ reconstructs $x$ by concatenating $c$ with $z$. 
Like VAE-GAN \cite{larsen2015autoencoding}, cVAE can also be extended to cVAE-GAN as introduced in \cite{bao2017cvae}. 

In cVAE, the label relevant and irrelevant factors are not explicitly disentangled in the posterior $q_\phi(z|x,c)$, hence manipulating $z\sim q_\phi(z|x,c)$ often leads to the unnecessary change on the given condition. 
Moreover, the synthesized images in cVAE often have the low conformity on the condition. 
To improve the conditional synthesized results, some works try to disentangle label relevant factors from all the others. CSVAE \cite{klys2018learning}  proposes to learn the conditional label relevant subspace by using distinct priors under the different labels. 
Zheng \emph{et. al.} \cite{zheng2019disentangling} employ two encoders. One learns the label dependent priors and specifies the posterior $z_s$ as a $\delta$ function. The other uses a $N(0, I)$ Gaussian prior, and maps the data into a label irrelevant posterior $q_\phi(z_u|x)$ . 
Both the works adopt the adversarial training in the latent space to prevent $z_u$ from carrying the label information.

Different from previous works, we use a single prior, regardless of the label of input, and model the label irrelevant posterior based on the encoder. A deterministic code 
is also employed to model label relevant factors. 
Moreover, we do not apply the adversarial training to minimize the mutual information between the label and its irrelevant factors, which makes the optimization easy. The key idea is to disentangle the label irrelevant code $z_u$, sampled from the posterior of VAE encoder, and label relevant code $z_s$, given by the condition mapping network, based on the proposed cVAE structure. We design the two types of structures.
They are chosen depending on whether the label condition is spatial related (\emph{e.g.} the pose degree) or not (\emph{e.g.} the face identity). Posteriors in both cases are constrained in the same way, close to the prior $N(z;0, I)$, thus the code $z_u\sim N(z;0, I)$ is label irrelevant. The dimension of $z_s$ and $z_u$ are intentionally designed so that they become complementary.  Particularly, one of them reflects the spatial structure, so it is a single channel feature map and 
is applied into VAE through spatially-adaptive normalization (SPADE) \cite{park2019semantic}. The other is a $1\times 1\times C $ style vector 
and it is incorporated into VAE by adaptive instance normalization (AdaIN) \cite{huang2017arbitrary,karras2019style}. Note that Fig.\ref{fig:fig21} shows one case in which $z_s$ is a style code and $z_u$ from the posterior is a one channel feature map. 
To improve the image quality, we add a discriminator like cVAE-GAN \cite{bao2017cvae}, and extend the fake data to include the condition exchanged image. Therefore, the reconstructed, prior-sampled and condition exchanged images are all regarded as fake ones for the discriminator.

Our contribution lies in following aspects. First, we propose a simple, flexible way to disentangle the spatial structure and style code for synthesis. It requires one of them to be label dependent, 
and it is given to VAE as the condition. The other 
becomes label irrelevant posterior after training. 
Second, by applying the adaptive normalization based on both the style and the spatial structure code, our model improves the disentangling performance. We carry out experiments on two types of datasets to prove the effectiveness of the method.

\section{Proposed Method}
\label{sec:Pro}
\begin{figure}[htb]
  \centering
  \includegraphics[width=0.5\textwidth]{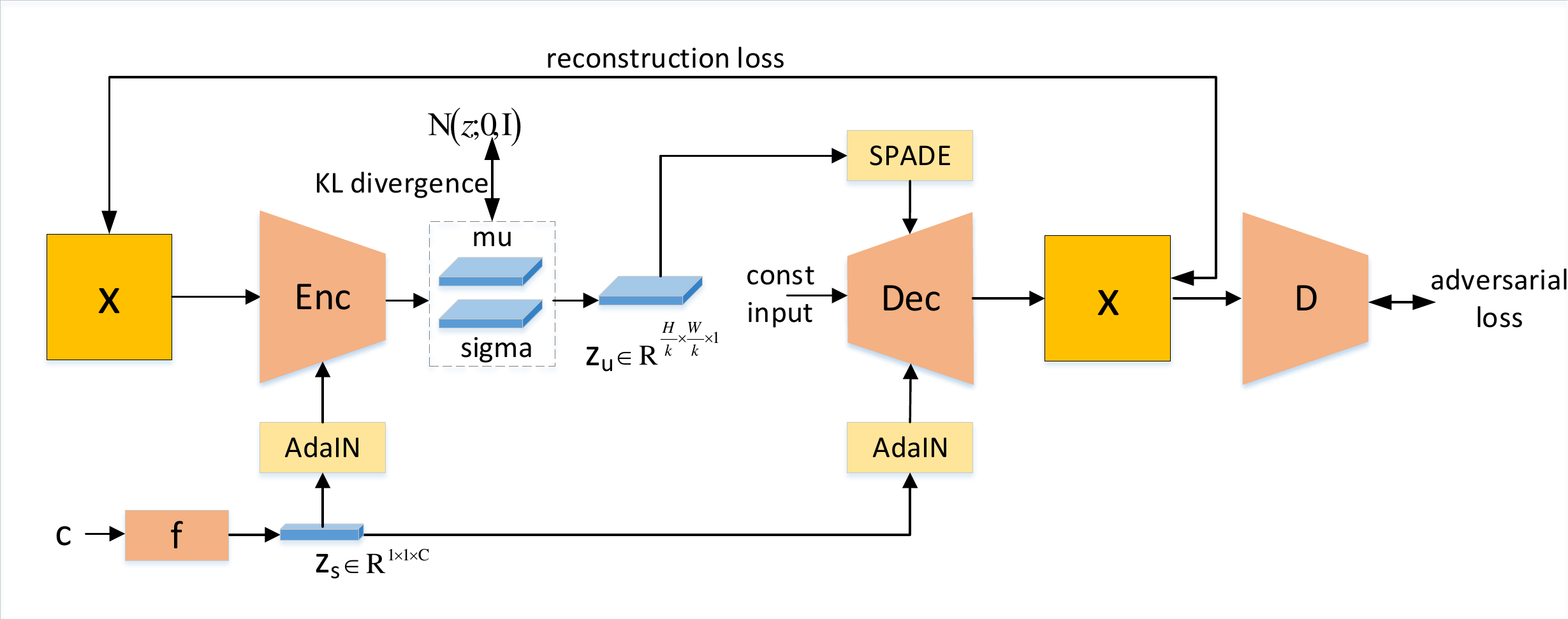}
  \vspace{-1.0cm}
  \caption{The proposed architecture on FaceSrub, in which ID labels have few spatial cues, and they are mapped into style codes. \emph{Dec} exploits the spatial structure code $z_u\in \mathbb{R}^{\frac{H}{k}\times\frac{W}{k}\times 1}$ with SPADE, and the style code $z_s\in\mathbb{R}^{1\times 1\times C}$ with AdaIN. 
  }
  \vspace{-0.5cm} 
  \label{fig:fig21}
\end{figure}

Fig.\ref{fig:fig21} shows the overall architecture. The generator $G$ is of VAE structure, consisting of the \emph{Enc}, \emph{Dec} and the condition mapping network $f$. \emph{Enc} specifies the spatial structure preserving posterior which is assumed to be label irrelevant, and constrained with prior $N(0, I)$ by KL divergence. \emph{Dec} exploits the spatial structure code $z_{u}$ sampled from the posterior with SPADE, and the style code $z_s$ given by $f$ with AdaIN. The "const input" of \emph{Dec} does not affect the appearance of the generated image, hence it can be a constant vector. Note that for brief, we do not distinguish the random variables and their realizations on $z_u$ and $z_s$. 
\vspace{-0.2cm} 

\subsection{cVAE formulation}
Supposing the image $x\in \mathbb{R}^{H\times W\times 3}$ with its label $c$, the goal of cVAE is to maximize the ELBO defined in the left of  (\ref{eq:eq1}), so that the data log-likelihood $\log p(x)$ can be maximized. Here $H$ and $W$ indicate the height and width of the input image, $\phi$, $\theta$ and $\psi$, correspond to the model parameters in \emph{Enc}, \emph{Dec} and $f$, respectively. 
The key idea is to split the latent $z$ into separate codes, the label relevant $z_s$ and irrelevant $z_u$. $D_{KL}$ indicates the KL divergence between two distributions. Note that there are three terms in (\ref{eq:eq1}). The first one is the negative reconstruction error. The second and third terms are the regularization which pushes the $q_{\phi, \psi}(z_u|x,c)$ and $q_\psi(z_s|c)$ to their priors $p(z_u)$ and $p(z_s)$, respectively. In practice, we assume that $z_s$ is deterministic, which means $p(z_s)$ and $q_\psi(z_s|c)$ are both dirac $\delta$ function. Hence the third term is strictly required to be $0$, thus can be ignored.
\
\begin{equation}\label{eq:eq1}
\begin{aligned}
    \log p(x)\geq \mathbb{E}_{q_{\phi, \psi}(z_u|x,c),q_\psi(z_s|c)}[\log p_\theta(x|z_u,z_s)]\\-D_{KL}(q_{\phi, \psi}(z_u|x,c)||p(z_u))-D_{KL}(q_\psi(z_s|c)||p(z_s))
\end{aligned}
\end{equation}

\subsection{Details about the network}
\subsubsection{Conditional label mapping network $f$}
As is illustrated in Fig.\ref{fig:fig21}, the input of $f$ is a conditional label $c$, which indicates the category of $x$, usually expressed in a one-hot vector. Like \cite{karras2019style}, we use several fully-connected layers to map $c$ into an embedding code $z_s$, which is later employed by the $Enc$ and $Dec$ based on the adaptive normalization module. Here $C$ is the number of channels, 
$k$ is the ratio of the height (or width) of the image to the feature map, $z_s=f(c)$ is the label relevant code 
, and it is reshaped either as a spatial structure preserving feature map $z_s\in \mathbb{R}^{\frac{H}{k}\times \frac{W}{k}\times 1}$, or a style code $z_s\in \mathbb{R}^{1\times 1\times C}$. We make the choice based on whether $c$ directly relates with the spatial structure. Actually, we try both cases on two different datasets, 3D chair\cite{aubry2014seeing} and FaceScrub\cite{ng2014data}. The details are given in the experiments section.

\vspace{-0.2cm}
\subsubsection{Encoder and Decoder}
cVAE has a pair of connected \emph{Enc} and \emph{Dec}. Together with $f$, \emph{Enc} takes the image and label pair $\{x,c\}$ and maps it into a posterior probability, which is assumed to be the Gaussian $q_{\phi,\psi}(z_u|x,c)=N(z_u|\mu,\sigma)$. Here $\mu$ and $\sigma$ are the mean and standard deviation, depended on $\{x,c\}$ and output from \emph{Enc}. A code $z_u\sim q_{\phi,\psi}(z_u|x,c)$ is given to $Dec$ to reconstruct $x$. Note that in VAE, there is a prior on $z_u$, which is $p(z_u)=N(z_u;0, I)$. During the optimization, the KL divergence between the posterior and the prior $D_{KL}(q_{\phi,\psi}(z_u|x,c)||p(z_u))$ is considered. In other words, \emph{Enc} maps $x$ from various classes into the posterior close to the same prior. Therefore $z_u$ becomes label irrelevant.

We have two choices on the shape and dimension settings for $z_u$ and $z_s$. 
Similar with $z_s$, 
$z_u$ is either a one channel feature map $z_u\in\mathbb{R}^{\frac{H}{k}\times \frac{W}{k}\times 1}$ when $z_s\in\mathbb{R}^{1\times 1\times C}$, or a style code $z_u\in\mathbb{R}^{1\times 1\times C}$ when $z_s\in\mathbb{R}^{\frac{H}{k}\times \frac{W}{k}\times 1}$. In the former, $z_u$ keeps the spatial structure from $x$, 
and in the latter, it is formulated to capture the feature channel style by the global average pooling. 
Section \ref{sec:exp} demonstrates that the dimension settings for $z_s$ and $z_u$ in the above two cases achieve the best performance on FaceScrub and 3D chair, respectively.

\emph{Dec} takes $z_s$ and $z_u$ to reconstruct $x$. Traditionally in cVAE, all inputs are directly concatenated along the channel at the beginning.  
However, this simple strategy does not emphasize the difference between $z_s$ and $z_u$, which definitely degrades the disentangling quality. Inspired by 
two adaptive normalization structures, AdaIN \cite{karras2019style} and SPADE \cite{park2019semantic}, we use them to help disentangling $z_s$ and $z_u$. 
As shown in (\ref{eq:eq2}), $h^{(l)}\in {\mathbb R^{N\times H^{(l)}\times W^{(l)} \times C^{(l)}}}$ is an intermediate output 
in $l$th layer of \emph{Dec}. $\hat{h}^{(l)}$ is the tensor after the normalization. Here, $H^{(l)}$, $W^{(l)}$ and $C^{(l)}$ indicate the height, width and channel number of $h^{(l)}$. $N$ is the batch size. $\gamma(z)$ and  $\beta(z)$ are two outputs from the conditional branch, depended on $z_s$ or $z_u$. $\mu_{h}$ and $\sigma_{h}$ are the mean and standard deviation statistics on $h^{(l)}$. SPADE and AdaIN have different strategies to 
compute $\mu_{h}$ and $\sigma_{h}$, and manipulate $\beta$ and $\gamma$. In SPADE, $\mu_{h}$ and $\sigma_{h}$ are computed over the dimensions of $H^{(l)}$, $W^{(l)}$ and $N$, while $\beta$, $\gamma$ are provided with distinct values along the 3D tensor. In AdaIN, $\mu_{h}$ and $\sigma_{h}$ are computed over the spatial dimensions, but $\beta$ and $\gamma$ are only provided for channel dimensions. Our method processes 
$h^{(l)}$ by both SPADE and AdaIN. Then we concatenate the results and reduce the channels by $1\times 1$ conv.
\vspace{-0.2cm}
\begin{equation}\label{eq:eq2}
{\hat{h}}^{(l)}=\gamma(z)\frac{{ h}^{(l)}-{\mu_{h}}}{\sigma_{h}}+\beta(z), z\in\{z_s, z_u\}
\end{equation}
\vspace{-0.8cm}

\subsubsection{Discriminator}
Traditional cVAE has only the \emph{Enc} and \emph{Dec}. Thus they are optimized only by the reconstruction loss with the KL divergence as a regularization like (\ref{eq:eq1}). To improve the synthesis quality, 
we add adversarial training by employing a discriminator $D$ 
to inspect the quality of fake data. 
Besides, we can have more types of fake data. 
In \cite{larsen2015autoencoding,bao2017cvae}, the reconstructed and prior sampled image from $z_u\sim N(z;0,I)$ are given to $D$. Our work extends it by synthesizing this kind of exchanged image by $z_s$ and $z_u$ from images of different conditions with labels of $c'$ and $c$. $D$ proposed in \cite{miyato2018cgans} is used in our work. Its score is from -1 to 1. The adversarial training loss 
is in (\ref{eq:eq3}).
\vspace{-0.2cm}
\begin{equation}\label{eq:eq3}
\begin{aligned}
    L_{D}^{adv} =& \mathbb{E}_{{x}\sim p_{\rm{data}}}[\max(0,1-D(x,c))]
    \\&+\mathbb{E}_{z_{u}\sim N(z; \mu,\sigma)}[\max(0,1+D(Dec(z_u, f(c)),c))]\\
 &+\mathbb{E}_{z_{u}\sim N(z; \mu, \sigma)}[\max(0,1+D(Dec(z_u, f(c')),c'))]\\
&+\mathbb{E}_{z_u\sim N(z; 0, I)}[\max(0,1+D(Dec(z_u, f(c)), c))]
\end{aligned}
\end{equation}
Here the first term is the score for real image, and the other three terms are the scores for the reconstructed, exchanged and prior sampled image.
\vspace{-0.1cm}
\section{Experiments}
\label{sec:exp}

\vspace{-0.2cm}
\subsection{Experimental setup}
\textbf{Datasets.} We conduct experiments on two datasets, including the 3D chair\cite{aubry2014seeing} and the FaceScrub\cite{ng2014data}. The 3D chair 
depicts a wide variety of chairs in 62 different chosen azimuth angles. Images are resized to the fixed size $64\times 64$.
The Facescrub contains $107$k facial images from 530 different IDs. 
These faces are cropped by the detectors \cite{chen2014joint}, and they are aligned 
based on the facial landmarks \cite{xiong2013supervised}. The detected cropped images are in $128\times 128$.

\textbf{Network structure and dimension settings} on $z_s$ and $z_u$. Table \ref{tbl:table31} lists the comparison structures in our experiments on two datasets. Typically, $z_s$ and $z_u$ is used by \emph{Dec} in 3 ways, ``\emph{Dec} inputs'', ``AdaIN'', and ``SPADE''. ``\emph{Dec} inputs'' indicates that the code is taken as \emph{Dec}'s input. 

On \textbf{3D Chair} dataset, the model rotates chair to the specified azimuth, while preserving original chair style. Here, the label relevant code $z_s$ has the spatial structure, and the label irrelevant $z_u$ is the style code. Different from Fig.\ref{fig:fig21}, \emph{Dec} of the proposed structure adopts $z_{u}$ with AdaIN, and $z_s$ with SPADE, and it is compared with other structures from S1 to S4. On \textbf{ Facescrub}, the model changes facial image to the specified ID but preserve label irrelevant factors, like pose or expression, from the input image. It is obvious that the ID relevant $z_s$ becomes the style code and the ID irrelevant $z_u$ mainly specifies spatial related information. The proposed structure, shown in Fig.\ref{fig:fig21}, is compared with S1 and S3, two typical structures using AdaIN. 
To compare different models in a fair way, we fix the total dimension of $z_s$ and $z_u$ to 512, and each is 256. If it is a spatial structure code, its size is $16\times 16\times 1$. While for style code, it is $1\times 1\times 256$. 


\textbf{Evaluation metrics.} We adopt three metrics for quantitative analysis. (1) Classification Accuracy (\emph{Acc}) reflects the condition conformity of the generated images. We use ResNet-50 \cite{he2016deep} trained on these two datasets for evaluating. 
(2) Fr$\Acute{e}$chet Inception Distance (\emph{FID})\cite{heusel2017gans} measures the distance between distributions of the synthesized and the real images, thus the lower, the better. (3) Mutual Information (\emph{MI}), between label irrelevant code $z_{u}$ and original label $c$.
\begin{equation}\label{eq:eq31}
\begin{split}
    MI=I(z_{u};c) 
 =\mathbb{E}_{q(z_{u}|c)p(c)}\log\frac{q(z_{u}|c)}{q(z_{u})}\\
=\frac{1}{N_C}\sum_{c} \mathbb{E}_{q(z_{u}|c)}\log\frac{q(z_{u}|c)}{q(z_{u})}
\end{split}
\end{equation}
\emph{MI} can be computed as (\ref{eq:eq31}), and the smaller, the better. It indicates whether label relevant and irrelevant variables are disentangled well. Here $N_C$ is the number of the categories. $q(z_{u}|c)$ and $q(z_{u})$ are approximated from the posterior distribution by Monte Carlo simulation.




\subsection{Results}
\begin{figure}[!htb]\small
  \centering
  \includegraphics[width=0.5\textwidth]{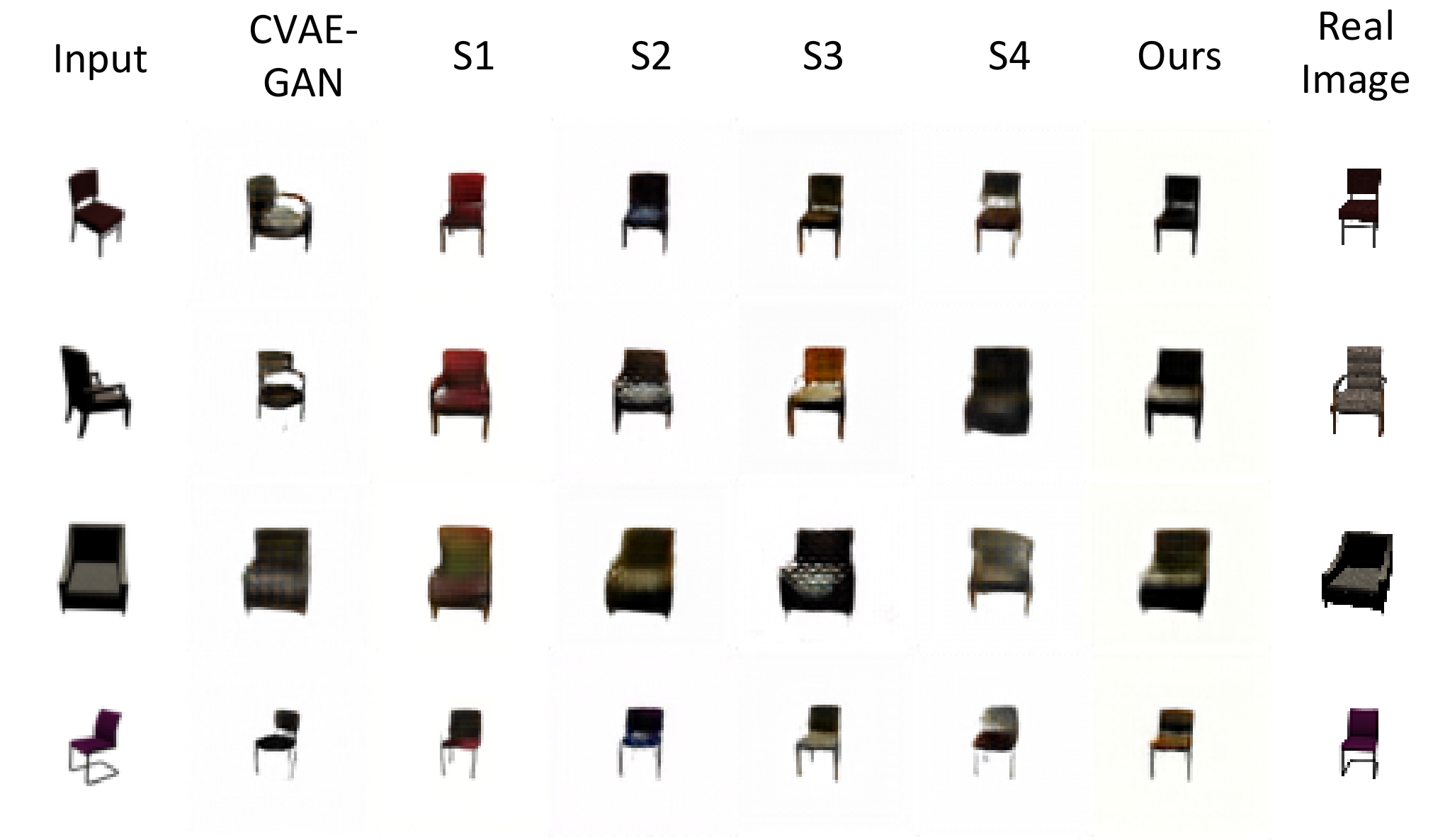}
  \caption{Rotated (exchanged) images comparison on 3D chair. }
  \label{fig:fig32}
\end{figure}

\begin{figure}[!htb]\small
  \centering
  \includegraphics[width=0.5\textwidth]{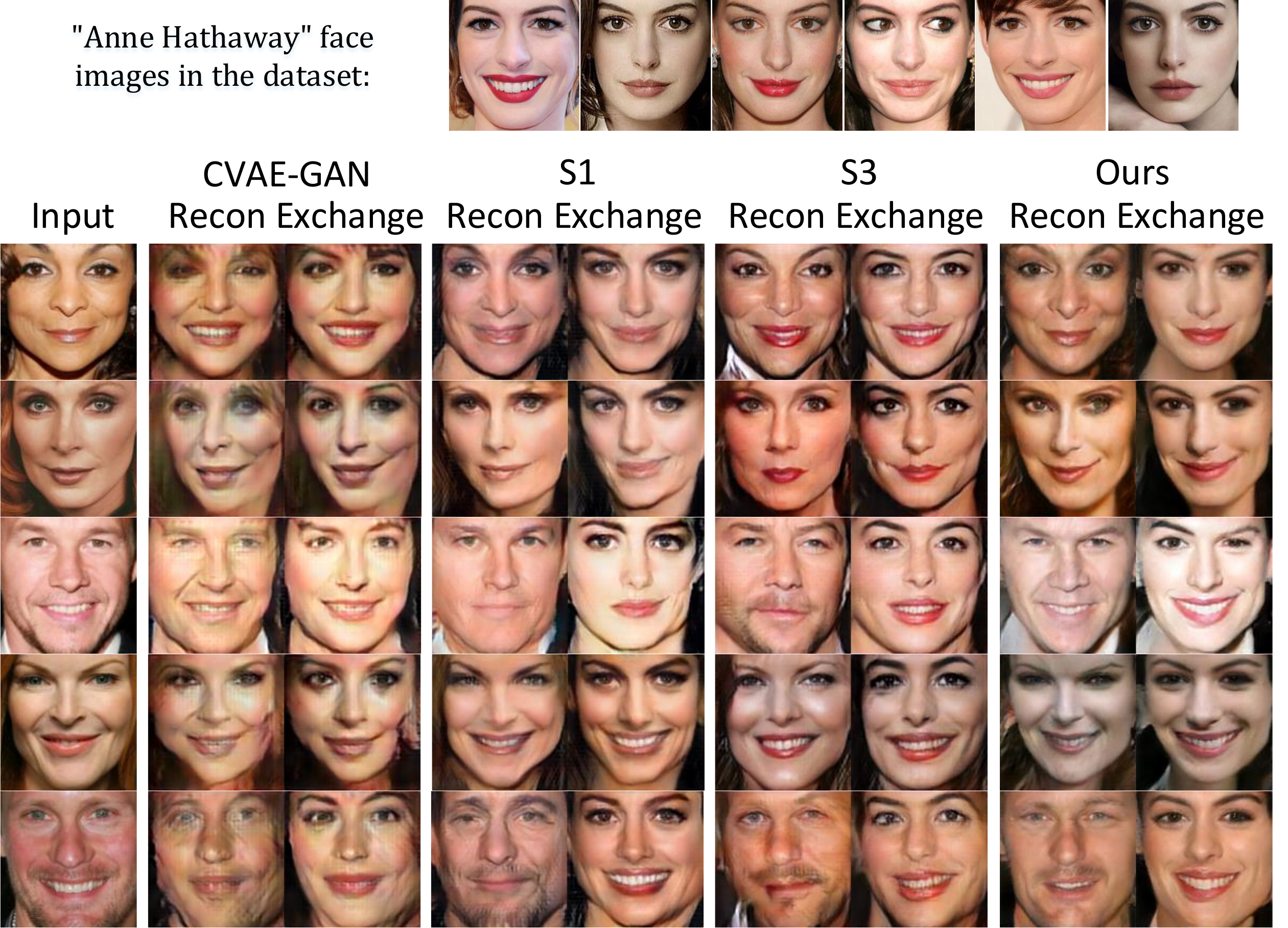}
  \caption{Reconstructed and exchanged images on FaceScrub.}
  \label{fig:fig33}
\end{figure}

\vspace{-0.2cm}
\textbf{Qualitative results.} We choose one specific label to synthesize the exchanged images under the condition. 
For 3D chair, the condition label is \emph{``24''}(which corresponds to azimuth $278^\circ$) and for Facescrub it is \emph{``Anne Hathaway''}. The results from different models on 3D chair and Facescrub are presented in Fig.\ref{fig:fig32} and Fig.\ref{fig:fig33}, respectively.  In both figures, images generated from our proposed method achieves the best performance. Particularly, the proposed model keeps the style of input chair well when rotating it, and it also maintains the facial expression when changing its identity on FaceScrub. 


\begin{table*}[htbp]\small
    \centering
    \caption{The compared structures on two datasets. 
    Note that we do not compare all possible ways. But only the typical ones. }
    \label{tbl:table31}
	\begin{tabular}{p{0.5cm}<{\centering}|p{1.8cm}<{\centering}|p{1.4cm}<{\centering}p{1.4cm}<{\centering}p{1.4cm}<{\centering}p{1.4cm}<{\centering}p{1.4cm}<{\centering}|p{1.4cm}<{\centering}p{1.4cm}<{\centering}p{1.4cm}<{\centering}}
    \toprule
     &   & \multicolumn{5}{c}{3D chair}                                    & \multicolumn{3}{c}{Facescrub}        \\ \midrule
     & cVAE-GAN  & S1 & S2 & S3 & S4 & Proposed & S1 & S3 & Proposed \\
$z_s$ & \emph{Dec} input    & AdaIN      & SPADE      & AdaIN        & AdaIN       & SPADE    & AdaIN      & AdaIN        & AdaIN    \\
$z_u$ & \emph{Dec} input & \emph{Dec} input  & \emph{Dec} input  & AdaIN        & SPADE       & AdaIN    & \emph{Dec} input  & AdaIN        & SPADE  \\ \bottomrule
\end{tabular}
\end{table*}

\textbf{Quantitative results.} To validate the proposed model, we compute the \emph{MI} and \emph{Acc} on 3D chair, and the \emph{Acc} and \emph{FID} on Facescrub. Results are shown in Table \ref{tbl:table32} and Table \ref{tbl:table33}. For \emph{MI} and \emph{Acc}, we generate 100 exchanged images for each specified label. For \emph{FID}, due to many IDs in Facsescrub, we choose to evaluate it on 5 IDs. 
Each ID has 5000 images and we calculate the average \emph{FID} on the chosen 5 categories. As shown in  Table \ref{tbl:table32} and Table \ref{tbl:table33}, our proposed method, which uses SPADE and AdaIN to process the spatial structure and style code respectively, achieves the best scores on both datasets.
\begin{table}[H]
    \centering
    \caption{\emph{MI} and \emph{Acc} on 3D chair from different models.}
    \label{tbl:table32}
	\begin{tabular}{p{2.5cm}<{\centering} p{1.8cm}<{\centering} p{1.8cm}<{\centering}}
    \toprule
         & \emph{MI} & \emph{Acc}   \\ \midrule
cVAE-GAN & 3.777 & 0.124 \\
S1 & 3.773 & 0.573 \\
S2 & 3.765 & 0.608 \\
S3 & 3.775 & 0.511 \\
S4 &  3.774   & 0.404 \\
Proposed method  & \bf{3.750} & \bf{0.623} \\ \bottomrule
    \end{tabular}
\end{table}
\vspace{-0.8cm}
\begin{table}[H]
    \centering
    \caption{\emph{Acc} and \emph{FID} on Facescrub from different models.}
    \label{tbl:table33}
	\begin{tabular}{p{2.5cm}<{\centering} p{1.8cm}<{\centering} p{1.8cm}<{\centering}}
    \toprule
         & \emph{Acc}  & \emph{FID}\\ \midrule
cVAE-GAN & 0.072  & 83.05 \\
S1   & 0.444 & 80.37 \\
S3   & 0.498 & 52.46 \\
Proposed method  & \bf{0.632} & \bf{50.14} \\ \bottomrule
    \end{tabular}
\end{table}


%
%
%


\section{Conclusion}
\label{sec:conc}
\vspace{-0.2cm}
We propose a latent space disentangling algorithm for conditional image synthesis in cVAE. Our method divides the latent code into label relevant and irrelevant parts. One of them preserves the spatial structure, and the other is the style code. These two types of codes are applied into cVAE by different adaptive normalization schemes. Together with a discriminator in the pixel domain, our model can generate high quality images, and achieve the disentangling performance. 
\vfill\pagebreak

\bibliographystyle{IEEEbib}
\small{\bibliography{strings,refs}}

\end{document}